\documentclass[10pt,twocolumn,letterpaper]{article}

\usepackage{iccv}              
\usepackage{times}
\usepackage{epsfig}
\usepackage{graphicx}
\usepackage{amsmath}
\usepackage{amssymb}
\usepackage{booktabs}
\usepackage{indentfirst}
\usepackage{multirow}
\usepackage{pifont}
\usepackage{comment}
\usepackage{subfig}
\usepackage{makecell}
\usepackage{fnpos}

\mathchardef\mhyphen="2D

\usepackage{color}
\usepackage{xcolor}
\usepackage{caption}
\usepackage{mmstyle}

\newcommand{\inc}[1]{{\color{blue}{\textbf{$\uparrow$ #1}}}}

\newcommand{\tablestyle}[2]{\setlength{\tabcolsep}{#1}\renewcommand{\arraystretch}{#2}\centering\footnotesize}
\newlength\savewidth\newcommand\shline{\noalign{\global\savewidth\arrayrulewidth
		\global\arrayrulewidth .8pt}\hline\noalign{\global\arrayrulewidth\savewidth}}

\usepackage[pagebackref=true,breaklinks=true,letterpaper=true,colorlinks,bookmarks=false]{hyperref}

\usepackage[capitalize]{cleveref}
\crefname{section}{Sec.}{Secs.}
\Crefname{section}{Section}{Sections}
\Crefname{table}{Table}{Tables}
\crefname{table}{Tab.}{Tabs.}

\iccvfinalcopy

\def\iccvPaperID{6583}

\begin{document}

\title{SkeleTR: Towrads Skeleton-based Action Recognition in the Wild }

\author{
    Haodong Duan$^1$\thanks{The work was done during an Amazon internship.}
    \hspace{1.1cm} Mingze Xu$^2$
    \hspace{1.1cm} Bing Shuai$^2$
    \hspace{1.1cm} Davide Modolo$^2$ \\ [.5ex]
    \hspace{1.1cm} Zhuowen Tu$^2$
    \hspace{1.1cm} Joseph Tighe$^2$
    \hspace{1.1cm} Alessandro Bergamo$^2$ \\ [.5ex]
    $^1$The Chinese University of Hong Kong \hspace{0.9cm}
    $^2$AWS AI Labs \\ [.5ex]
    {\tt\small dhd.efz@gmail.com, \{xumingze,bshuai,dmodolo,ztu,tighej,bergamo\}@amazon.com}
}

\maketitle

\begin{abstract}

We present SkeleTR, a new framework for skeleton-based action recognition.
In contrast to prior work, which focuses mainly on controlled environments, 
we target more general scenarios that typically involve a variable number of people and various forms of interaction between people. 
SkeleTR works with a two-stage paradigm.
It first models the intra-person skeleton dynamics for each skeleton sequence with graph convolutions, 
and then uses stacked Transformer encoders to capture person interactions that are important for action recognition in general scenarios.
To mitigate the negative impact of inaccurate skeleton associations, 
SkeleTR takes relative short skeleton sequences as input and increases the number of sequences.
As a unified solution,
SkeleTR can be directly applied to multiple skeleton-based action tasks, 
including video-level action classification, instance-level action detection, and group-level activity recognition.
It also enables transfer learning and joint training across different action tasks and datasets, which result in performance improvement.
When evaluated on various skeleton-based action recognition benchmarks, 
SkeleTR achieves the state-of-the-art performance.

\end{abstract}
    
\section{Introduction}

Skeleton-based action recognition has received increasing attention in recent years due to its robustness against background and illumination changes~\cite{yan2018spatial,duan2022revisiting}.
Existing methods~\cite{liu2020disentangling,chen2021channel,chi2022infogcn,duan2022pyskl} mostly use Graph Convolutional Networks (GCNs) as backbones and focus on modeling spatio-temporal information from long skeleton sequences separately.
While observing promising results in simplified and controlled environments~\cite{shahroudy2016ntu,liu2020ntu},
they are less effective in more realistic scenarios with a variable number of people performing diverse actions.
First, considering that linking skeletons (\eg, using skeletons' confidence scores~\cite{yan2018spatial}) in a long time span may inevitably generate more association errors~\cite{dendorfer2020mot20,shuai2022large},
requiring long sequences as input can bring lots of noises.
Second, prior works pay little attention to modeling the relations among a group of people, which plays an important role in recognizing both person interactions (\eg, talking to someone~\cite{gu2018ava}) and group activities (\eg, pass/win-point in a volleyball game~\cite{ibrahim2016hierarchical}).

\begin{figure}[t]
	\centering
	\captionsetup{font=small}
	\includegraphics[width=\columnwidth]{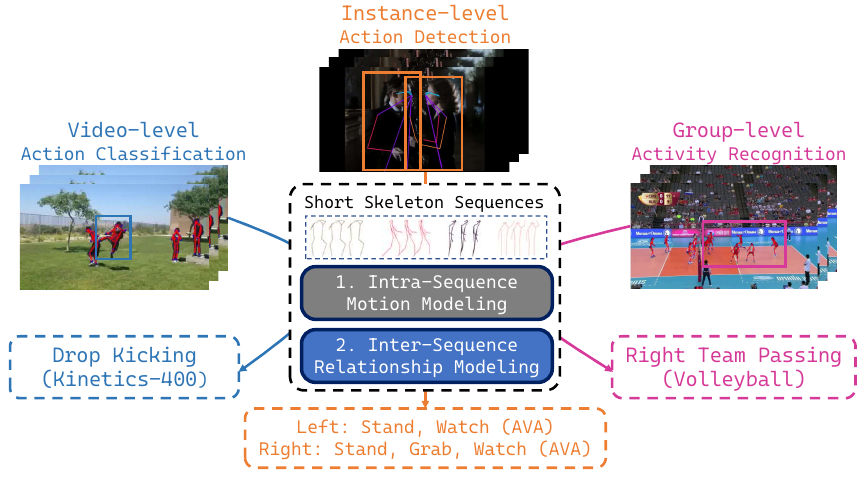}
	\vspace{-3mm}
	\caption{\textbf{SkeleTR is a general framework for skeleton action recognition which can handle different tasks}, including video-level action classification, instance-level action detection, and group-level activity recognition.
	It adopts a two-stage paradigm: the intra-sequence and inter-sequence modeling, respectively.}
	\label{fig-teaser}
    \vspace{-5mm}
\end{figure}

In this paper, we propose SkeleTR for skeleton action recognition in real-world scenarios and handle multiple action tasks (\ie, video-level, instance-level, and group-level action recognition) in a general framework.
SkeleTR samples a large number of short skeleton sequences instead of a few long ones (practice of prior works~\cite{yan2018spatial,duan2022revisiting}) from each video.
Each short sequence directly corresponds to an atomic action of a single person, 
thus the collection of multiple short sequences in space and time can represent hierarchical actions (with a longer time span) of a single person or a video-level action involving multiple people. 
The new input paradigm can be applied to instance-level and video-level recognition of actions with different temporal granularities,
and it has several unique advantages.
First, association errors are less likely to happen in a shorter skeleton sequence~\cite{dendorfer2020mot20,shuai2022large}. 
We show that the improved sequence consistency leads to better intra-sequence modeling.
Second, this new format is more flexible and helps to create comprehensive yet compact inputs. 
With a finer temporal granularity, the same input size budget can cover more human skeletons. 
It reduces redundant zero-paddings for a small number of people,
and mitigates information loss during input sampling for a large number of people.

SkeleTR adopts a two-stage paradigm as shown in Fig~\ref{fig-teaser}. 
Stage 1 performs intra-sequence modeling to capture the motion patterns of each individual skeleton sequence with a GCN backbone.
The GCN backbone outputs a spatio-temporal feature for each sequence.
Stage 2 concatenates all sequence features and feeds them into Transformer encoders to model inter-sequence relationships.
At the end of stage 2, average pooling aggregates all features belonging to the same sequence into a single feature vector.
By leveraging inter-sequence modeling,
the output sequence feature incorporates the interactions between people and the global context of the video, 
which is helpful in both instance-level and video-level action recognition.
To reduce the dimension of spatio-temporal skeleton features and keep the computations of Transformer tractable, 
we insert a novel Mix Pooling module before stage 2. 
It generates a compact representation with fine granularity for each sequence by applying multiple partial dimension reduction strategies in parallel.

SkeleTR easily enables transfer learning and joint learning across different datasets and tasks.
With a unified input format and model architecture, 
one can jointly train multiple tasks with different datasets, 
each with its own loss function and ground truth annotations.
This allows for leveraging auxiliary datasets for pre-training or joint training of a target task with limited labeled data.
In experiments, we find that joint training serves as a strong regularizer and improves recognition performance, especially in scenarios with limited training data.

To evaluate the effectiveness of SkeleTR, 
we conduct comprehensive study on three skeleton-based action recognition tasks:
action classification~\cite{carreira2017quo,yang2021unikau,shahroudy2016ntu}, spatio-temporal action detection~\cite{gu2018ava}, and group activity recognition~\cite{ibrahim2016hierarchical}.
Experimental results show that SkeleTR works well with different GCN backbones and significantly outperforms prior works on all three tasks.
For example, with ST-GCN++~\cite{duan2022pyskl} backbone, 
SkeleTR outperforms the state-of-the-art methods by 3.8\%, 3.8\%, 0.7\%, and 1.1\% on Posetics~\cite{yan2018spatial}, AVA~\cite{gu2018ava}, Volleyball~\cite{ibrahim2016hierarchical}, and NTU-Inter~\cite{pang2022igformer}.
With Posetics joint training,
the AVA performance is further improved by 4\% mAP.
The great capability of SkeleTR leads to another noteworthy result:
we first observe the strong complementarity of skeleton action recognition. 
Combining SkeleTR prediction and the RGB state-of-the-art~\cite{bertasius2021space,pan2021actor} with simple late fusion leads to great performance boost (2.3\% Posetics Top-1, 3\% AVA mAP).

\section{Related Works}

\noindent\textbf{Human Action Understanding.}
Action classification~\cite{carreira2017quo,goyal2017something} is the most fundamental task in human action understanding,
aiming at recognizing the video-level action in a short video clip. 
Another task of great interest is recognizing the instance/person-level actions in a video, 
either at the atomic~\cite{gu2018ava,li2020ava} or hierarchical~\cite{li2021multisports} level.
Meanwhile, other areas of interest include group activity recognition~\cite{ibrahim2016hierarchical,choi2009they}, action localization~\cite{caba2015activitynet,idrees2017thumos,zhao2019hacs}, or action segmentation~\cite{kuehne2014language,stein2013combining}. 
Various modalities in videos can be utilized to solve the action understanding tasks, such as 
RGB appearance~\cite{wang2016temporal,carreira2017quo,feichtenhofer2019slowfast,wang2018non}, 
optical flow~\cite{simonyan2014two}, audio waves~\cite{xiao2020audiovisual}, and human skeletons~\cite{yan2018spatial,duan2022revisiting}.
Spatio-temporal representations learned end-to-end from RGB appearance yield good performance on standard benchmarks. 
However, they are data intensive and less robust to context changes.
In this work, we propose a unified framework to learn video representations based on human skeletons, that can be easily applied to multiple action tasks.

\begin{figure*}[t]
	\centering
	\captionsetup{font=small}
	\includegraphics[width=\linewidth]{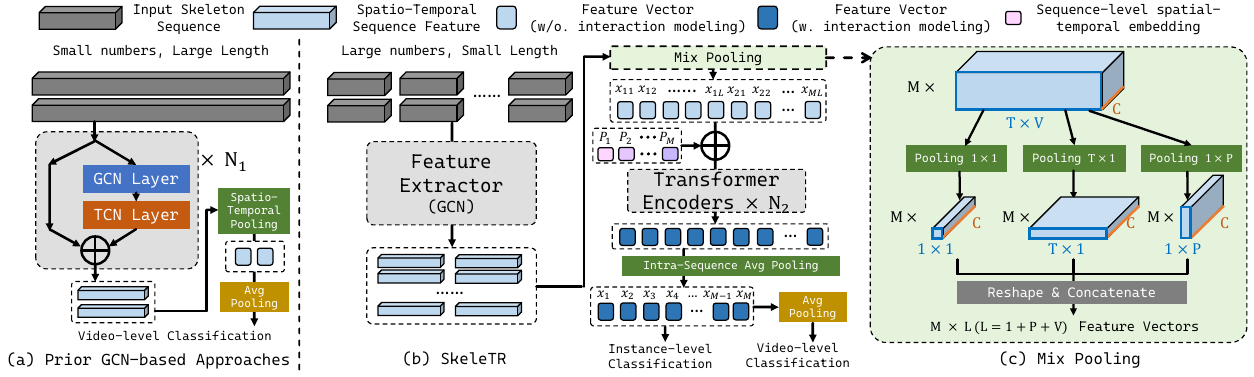}
	\vspace{-5mm}
	\caption{\textbf{The architecture of SkeleTR. } 
	SkeleTR takes a large set of short skeleton sequences as input and performs a two-stage modeling process.
	It first adopts a GCN backbone to model the intra-sequence skeleton dynamics. 
	Then it processes the sequence features with stacked Transformer encoders, 
	which helps to model person interactions and leverage the long-range context. 
	Between the two stages, we adopt a Mix Pooling layer to compress the spatio-temporal features and keep computations tractable. }
	\label{fig-arch}
	\vspace{-5mm}
\end{figure*}

\noindent\textbf{Skeleton-based Action Recognition.}
Human skeletons are robust against background and illumination changes~\cite{yan2018spatial,shi2019two}, 
and can be easily obtained with sensors~\cite{zhang2012microsoft} or pose estimators~\cite{8765346,sun2019deep}.
In skeleton action recognition, the majority of research has focused on modeling the intra-person skeleton dynamics. 
With the rise of deep learning, 
manual feature engineering~\cite{hussein2013human,vemulapalli2014human,wang2012mining} has gradually been replaced by end-to-end representation learning with deep neural architectures 
(GCN~\cite{yan2018spatial}, 2D-CNN~\cite{choutas2018potion}, or 3D-CNN~\cite{duan2022revisiting}). 
Among them, GCN-based approaches~\cite{yan2018spatial,shi2019two,liu2020disentangling,chen2021channel} have been the most popular paradigm, 
which construct spatial-temporal graphs of human keypoint coordinates and perform spatial and temporal modeling. 
These methods model the intra-person skeleton dynamics for in-lab collected videos~\cite{shahroudy2016ntu,liu2017pku}, featuring small number of persons (1 or 2) per video and accurate person identities. 
For more challenging scenarios~\cite{carreira2017quo}, 
the model performance decreases due to the noises produced in matching skeletons to sequences. 
Besides, GCN approaches primarily focus on predicting video-level actions, 
and do not benefit from context or interaction modeling when applied to instance-level action recognition. 
To this end, we propose SkeleTR, which expands existing GCN approaches and works universally well for diversified scenarios and recognition tasks.

\noindent\textbf{Human Interaction Recognition. }
Human interaction recognition~\cite{raptis2013poselet,vahdat2011discriminative,zhang2012spatio} is a sub-task of action recognition. 
It aims at identifying and categorizing activities between people.
Early works~\cite{yun2012two,ji2014interactive} built hand-crafted features for interaction modeling and adopted classic machine learning algorithms for activity classification.  
Subsequent works proposed to use two-stream RNNs to model the two-person interactions~\cite{perez2021interaction,men2021two,wang2017modeling}. 
Recently, IGFormer~\cite{pang2022igformer} first proposed to use a two-stream Transformer for two-person interaction modeling and observed improved performance.
However, most deep methods share the same limitations:
1. they are largely restricted to two-person scenarios and tend to suffer from exponentially growing inference costs with the person number;
2. they fail to leverage the strong intra-sequence modeling capabilities of GCNs.
To overcome these challenges, we introduce SkeleTR, which integrates intra-sequence and inter-sequence modeling with good scalability and small computation cost.

\noindent\textbf{Visual Transformer.}
With the great success in NLP~\cite{vaswani2017attention,devlin2018bert}, 
Transformers have been increasingly applied to diversified vision tasks~\cite{dosovitskiy2020image,liu2021swin,carion2020end,zhu2020deformable}
as well as the RGB-based action understanding~\cite{liu2022video,fan2021multiscale,arnab2021vivit,bertasius2021space,xu2021long,zhao2022tuber},
and have achieved promising results.
For skeleton action recognition, 
there exist works~\cite{plizzari2021skeleton,zhang2021stst} that directly use Transformers for intra-sequence spatio-temporal modeling. 
However, without exploiting the locality, the Transformer-based motion modeling only leads to inferior recognition performance compared to GCN-based ones.
In SkeleTR, to enjoy the best of both worlds, we adopt GCNs for intra-sequence motion modeling, 
and use Transformer encoders for inter-sequence relationship modeling among multiple persons' fine-grained skeleton features.

\section{SkeleTR}

\subsection{Overview}
Both SkeleTR and prior GCN approaches directly take human joint coordinates as input.
A person in a specific frame corresponds to a skeleton with $V$ joints (each has $C$ dimensions, $C$ is 2 or 3).
Skeletons in $T$ different frames (1 skeleton per frame) can be linked in temporal order to form a skeleton sequence $\mX_i \in \mathbb{R}^{T\times V\times C}$ with length $T$. 
A skeleton action recognizer takes $M$ skeleton sequences as input, and outputs a score vector $\mS \in \mathbb{R}^K$ for video-level action recognition, where $K$ is the number of action classes. 

Spatial-temporal GCNs are widely used to model skeleton sequences for action classification, 
which focus on modeling intra-sequence skeleton dynamics.
Most GCN methods~\cite{yan2018spatial,shi2019two,chen2021channel} share a similar high-level design (Fig~\ref{fig-arch}(a)),
where a backbone with $\mathrm{N_1}$ GCN blocks is used to extract features of each skeleton sequence. 
A GCN block comprises a GCN layer and a TCN layer. 
The GCN layer performs spatial modeling by fusing joint features within the same skeleton. 
The TCN layer employs 1D Convs to capture the motion patterns of each joint. 
GCNs model long skeleton sequences that span the entire video in parallel, 
and aggregate multiple sequence features with average pooling to predict the video-level action. 
On existing skeleton action recognition benchmarks, 
GCNs exhibit superior intra-sequence modeling capability than other architectures (\eg, RNN~\cite{zhu2016co,liu2016spatio} or Transformers~\cite{plizzari2021skeleton,zhang2021stst}).

SkeleTR inherits the effective intra-sequence modeling of GCNs, 
while also enhancing the flexibility, generality, and inter-person modeling capability.
SkeleTR takes $M$ short skeleton sequences as input (Sec.~\ref{sec-inputs}), and outputs one feature vector per sequence.
It first adopts a GCN backbone to model the motion patterns of each skeleton sequence,
then uses stacked Transformer encoders to capture the interactions and leverage long-range context (Sec.~\ref{sec:inter-seq}).
Between two stages, we use a Mix Pooling layer to compress spatio-temporal sequence features into a compact multi-view representation to keep computation tractable.
Fig~\ref{fig-arch} compares the design of SkeleTR and previous GCN methods. 

\begin{figure}[t]
	\centering
	\captionsetup{font=small}
	\includegraphics[width=\columnwidth]{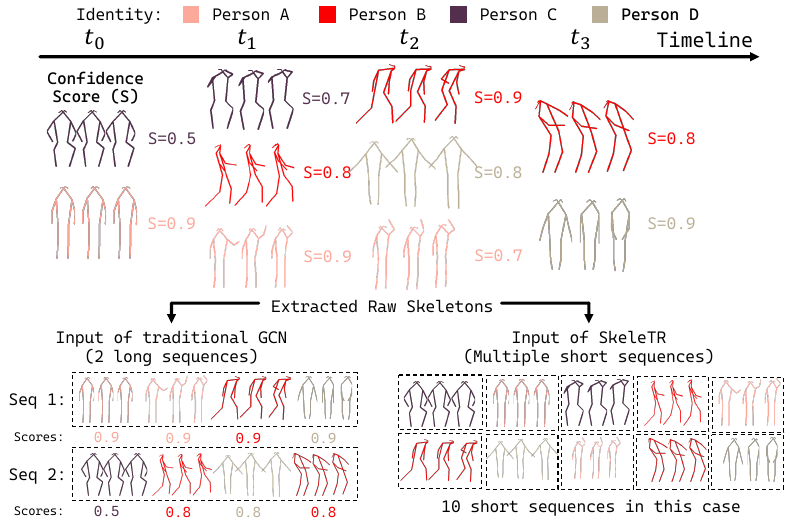}
	\vspace{-3mm}
	\caption{\textbf{The difference of input formats. } Prior works link skeletons at different timestamps based on the confidence score rank to form $\hat{M}=2$ long sequences, each covers the entire video. SkeleTR, instead, directly takes multiple shorter sequences as input. Each sequence only covers 1$\sim$2 seconds, thus has better consistency than long sequences. }
	\vspace{-5mm}
	\label{fig-input}
\end{figure}

\subsection{The Input Construction of SkeleTR}
\label{sec-inputs}

For skeleton action classification in real-world scenarios, 
prior works~\cite{yan2018spatial,shi2019two,liu2020disentangling} use a few long skeleton sequences as input, 
where skeletons at different timestamps are linked with rudimentary association methods.
They use a naive confidence score based matching to link skeletons to sequences as long as the entire video (as shown in Fig~\ref{fig-input}).
Specifically, they mostly keep $\hat{M}=2$ skeletons with the highest confidence scores per frame,
and link skeletons based on confidence rank. 
In cases where it is applied to challenging videos involving multiple persons, this strategy can lead to noisy skeleton sequences and sub-optimal performance. 

In contrast, we propose to sample a large number of short skeleton sequences as input, 
and link skeletons at different timestamps into sequences based on their IOU similarity. 
Specifically, to sample sequences with length $T$,
we first uniformly select a set of key frames with a stride of $T / 2$.
For each skeleton in every key frame, 
we create a sequence by linking skeletons within a window of $T$ frames using IoU-based matching~\cite{1517Bochinski2017}, 
resulting in a total of $M'$ sequences ($M'$ is variable for different videos).
Our new input format is compatible with a variable number of input sequences. 
By using different number of sequences ($M$) for each training batch, 
SkeleTR can handle inputs with different $M$, 
which helps to reduce the usages of zero-padding at test time and improves efficiency.

\subsection{Inter-sequence Modeling Stage}
\label{sec:inter-seq}

After processing short skeleton sequences with the GCN backbone, 
SkeleTR employs stacked Transformer encoders to model inter-sequence relationships. 
The process starts with reducing the feature dimension using a Mix Pooling layer, 
resulting in $L$ feature vectors for each sequence.
Next, a sequence-level spatial-temporal embedding is combined with the feature vectors. 
The resulting sequence features ($M\times L$ vectors in total) are concatenated and modeled using a standard Transformer with multiple encoder blocks~\cite{dosovitskiy2020image}. 
This architecture is depicted in Fig~\ref{fig-arch}(b). 
We present more details in the next paragraphs.

\noindent
\textbf{Mix Pooling. } 
The cost of computing the attention matrices in a Transformer block grows quadratically with the input dimensionality, 
resulting in a computational cost of $\Theta((M \times T \times V)^2)$ in our case, if no dimension reduction is performed ($L=T\times V$).
Given a video with a typical configuration ($T=15, V=17$) and a Transformer consisting of only $\mathrm{N_2} = 2$ encoder blocks, 
the Transformer itself consumes 52.7 GFLOPs, which is impractical to compute.
Another extreme is to remove the dimension $T$ and $V$ with global average pooling,
which reduces the computational complexity to $\Theta(M^2)$ but also results in loss of fine-grained features for each skeleton sequence.

Between two extremes ($L = T \times V$ or $L = 1\times 1$),
we adopt multi-stream partial dimension reductions to reduce the computation cost.
We denote a partial dimension reduction stream as $A \times B$, 
indicating reducing temporal dimension ($T$) to $A$ and spatial dimension ($V$) to $B$.
Each stream provides a specific view of the spatio-temporal skeleton features. 
Multiple reduction streams can be combined in Mix Pooling to compress raw features into a comprehensive representation. 
Specifically, we adopt 3 partial dimension reduction streams: $1 \times 1$, $T \times 1$, and $1 \times P$ ($V$ joints aggregated to $P=5$ parts: head, two arms, two legs) (Fig~\ref{fig-arch}(c)).
Mix Pooling concatenates the output of 3 streams and keeps $L=1+T+P$ features per sequence.
The heterogeneous features are then processed with one single Transformer.
Mix Pooling enables the cross-view feature interaction and helps keep computations tractable, yielding a complexity of $\Theta(M^2(1+T+P)^2)$ ($\sim 0.6$ GFLOPs / video). 

\noindent
\textbf{Sequence-level Spatial Temporal Embedding. }
We further combine $L$ features per sequence with a sequence-level spatial-temporal embedding. 
We utilize a fully-connected layer to encode both the bounding box and the normalized timestamp (computed based on the entire video) of the center skeleton for each sequence, which then serves as the embedding.
The embeddings are added with the Mix-Pooled features, 
and further processed by a layer normalization.

\noindent
\textbf{Transformer-based Interaction Modeling. } 
We use a standard Transformer with $\mathrm{N_2}$ encoder blocks for modeling sequence features. 
Each block consists of a multi-head self-attention layer (MHSA) and a feed-forward network (FFN), 
and adopts the post-layernorm style.
The multi-head design in the encoder blocks helps to capture various kinds of relationships among skeleton features, 
including but not limited to the inter-person interactions and intra-person long-range context. 
At last, we use intra-sequence average pooling to aggregate $L=(1+T+P)$ features per sequence into a single one. 
The output sequence features, which incorporate both person interactions and a comprehensive context, 
are then used to predict the instance-level or video-level action.
We present the detailed configuration of encoder blocks in the appendix.

\begin{figure}[t]
	\centering
	\captionsetup{font=small}
	\includegraphics[width=\columnwidth]{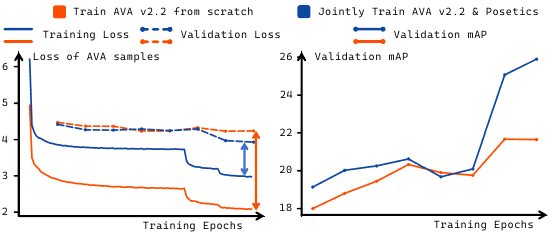}
	\vspace{-5mm}
	\caption{\textbf{The curve of training / validation loss and validation mAP on AVA.} 
	Joint training with Posetics~\cite{yang2021unikau} significantly alleviates overfitting for AVA action detection. 
	The jointly-trained model achieves much higher validation mAP, 
	with the gap of training / validation losses largely shrinks. } 
	\label{fig-curve}
	\vspace{-5mm}
\end{figure}

\subsection{SkeleTR as a General Framework}

As a general framework for skeleton action recognition, 
SkeleTR can be applied to various skeleton-based action recognition tasks.
It also enables and benefits from transfer learning and joint learning across different tasks. 

\noindent
\textbf{Video-level Action Classification.}
The tasks of action classification~\cite{carreira2017quo,yang2021unikau} and group activity recognition~\cite{ibrahim2016hierarchical,choi2009they} both take as input a video and predict the action at video-level.
Note that the input videos may contain a variable number of people. 
In SkeleTR, we sample numerous short skeleton sequences that last $1\sim2$ seconds.
We then average the $M$ output features, and use a fully-connected layer to predict the video action class.
Thanks to the local motion modeling with GCN and interaction modeling with Transformers, 
SkeleTR achieves strong performance in both action classification and group activity recognition. 

\noindent
\textbf{Instance-level Action Recognition.}
Spatio-temporal action detection~\cite{gu2018ava} aims to recognize actions of each person at specific timestamps (instance-level recognition). 
For this task, we sample short skeleton sequences with length $T$\footnote{
$T$ is restricted by annotation granularity. 
For example, AVA~\cite{gu2018ava} annotates atomic actions at 1fps, for which $T$ should cover at most 2 seconds.} 
in a longer temporal window ($5\sim 10$s) as input, and utilize the sequence-level features for instance-level action recognition.
SkeleTR enables the modeling of both intra-sequence motion patterns and inter-sequence interactions for skeleton-based spatio-temporal action detection. 

\noindent
\textbf{Transfer Learning and Joint Learning.} 
SkeleTR enables to:
1. transfer the representation learned on one task to another new task;
2. jointly train different datasets and tasks, which in turn alleviate the overfitting issues on small datasets.
Fig~\ref{fig-curve} shows the curves of training and validation losses, and validation mAP for AVA.
We see that joint training serves as a strong regularizer and alleviates the severe overfitting.
With joint training, the gap between training and validation losses largely shrinks, 
and the recognition performance on the validation set becomes much stronger. 

\section{Experiments}

\begin{table*}[t]
	\centering
	\captionsetup{font=small, position=bottom}
	\captionsetup[subfloat]{font=small, position=bottom}
    \resizebox{\linewidth}{!}{
	\subfloat[\textbf{Action Classification: Posetics.}  \label{tab-sota-pos}]{
		\tablestyle{4pt}{1.2}
        \begin{tabular}[t]{ccc}
        \shline
        Methods & \makecell{Posetics\\Top-1} & \makecell{Posetics\\Top-5} \\ \shline
        ST-TR~\cite{plizzari2021skeleton} & 47.5 & 71.3 \\
        MS-G3D~\cite{liu2020disentangling}  & 52.6 & 75.8 \\ 
        CTR-GCN~\cite{chen2021channel} & 54.1 & 76.7 \\ 
        ST-GCN++~\cite{duan2022pyskl} & 52.5 & 76.0 \\ 
        UNIK~\cite{yang2021unikau} & 52.5 & 75.7 \\
        PoseC3D~\cite{duan2022revisiting} & 53.1 & 77.1 \\ \shline
        SkeleTR-S & 55.9 & 78.9 \\ 
        SkeleTR-L & \textbf{57.9} & \textbf{80.3} \\ 
        SkeleTR [C] & 57.6 & 80.1 \\ \shline
        \end{tabular}}
	\hspace{2mm}
    \subfloat[\textbf{Action Recognition: NTU-Inter. } \label{tab-sota-ntu}]{
		\tablestyle{4pt}{1.2}
        \begin{tabular}[t]{ccccc}
        \shline
        Methods & \makecell{NTU60\\XSub} & \makecell{NTU60\\XView} & \makecell{NTU120\\XSub} & \makecell{NTU120\\XSet} \\ \shline
        ST-LSTM~\cite{liu2016spatio} & 83.0 & 87.3 & 63.0 & 66.6 \\ 
        LSTM-IRN~\cite{perez2021interaction} & 90.5 & 93.5 & 77.7 & 79.6 \\  
        GeomNet~\cite{nguyen2021geomnet} & 93.6 & 96.3 & 86.5 & 87.6 \\ 
        IGFormer~\cite{pang2022igformer} & 93.6 & 96.5 & 85.4 & 86.5 \\  \shline
        AS-GCN~\cite{li2019actional} & 87.1 & 92.0 & 77.8 & 79.3 \\ 
        CTR-GCN~\cite{chen2021channel} & 91.6 & 94.3 & 83.2 & 84.4 \\ \shline
        SkeleTR-S & 94.6 & 97.1 & 87.4 & 88.0 \\ 
        SkeleTR-L & 94.7 & 97.5 & \textbf{87.8} & \textbf{88.3} \\ 
        SkeleTR [C] & \textbf{94.8} & \textbf{97.7} & \textbf{87.8} & \textbf{88.3} \\ \shline
        \end{tabular}}
    \hspace{2mm}
	\subfloat[\textbf{Spatio-temporal Action Detection: AVA v2.2.} \label{tab-sota-ava}]{
		\tablestyle{4pt}{1.2}
        \begin{tabular}[t]{ccccc}
        \shline
        Methods & \makecell{Overall\\mAP} & \makecell{Person\\Movement} & \makecell{Object\\Manipulation} & \makecell{Person\\Interaction} \\ \shline
        ST-TR~\cite{plizzari2021skeleton} & 17.3 & 37.3 & 8.4 & 18.8 \\
        MS-G3D~\cite{liu2020disentangling} & 18.4 & 38.7 & 9.6 & 19.7 \\ 
        CTR-GCN~\cite{chen2021channel} & 18.9 & 39.4 & 9.8 & 20.5 \\ 
        ST-GCN++~\cite{duan2022pyskl} & 18.6 & 39.1 & 9.7 & 20.0 \\ 
        UNIK~\cite{yang2021unikau} & 19.0 & 39.7 & 9.9 & 20.4 \\ 
        PoseC3D~\cite{duan2022revisiting} & 18.3 & 39.3 & 9.3 & 19.1 \\  \shline
        SkeleTR-S & 22.8 & 43.8 & 12.3 & 26.9 \\ 
        SkeleTR-S* & 26.8 & 48.2 & \textbf{17.7} & 27.8 \\ 
        SkeleTR [C]* & \textbf{27.3} & \textbf{50.9} & 17.3 & \textbf{28.3} \\ \shline
        \end{tabular}}
    \hspace{2mm}
    \subfloat[\textbf{Group Activity Recognition: Volleyball. } \label{tab-sota-vol}]{
        \tablestyle{4pt}{1.2}
        \begin{tabular}[t]{cc}
        \shline
        Methods & \makecell{Volleyball\\Top-1} \\ \shline
        AT$^+$~\cite{gavrilyuk2020actor} & 92.3 \\ 
        POGARS~\cite{thilakarathne2022pose} & 93.2 \\ 
        COMPOSER~\cite{zhou2021composer} & 93.7 \\ \shline
        ST-TR~\cite{plizzari2021skeleton} & 93.3 \\
        CTR-GCN~\cite{chen2021channel} & 93.6 \\ 
        ST-GCN++~\cite{duan2022pyskl} & 93.3 \\  \shline
        SkeleTR-S & \textbf{94.4} \\ 
        SkeleTR [C] & \textbf{94.4} \\ \shline
        \end{tabular}}}
    \vspace{-1mm}
    \caption{\textbf{Comparing SkeleTR with previous state-of-the-art on versatile skeleton action recognition tasks.} [C] denotes CTR-GCN backbone; *\,denotes joint training with Posetics. 
    For Posetics and AVA, we follow the common practice~\cite{chen2021channel,yang2021unikau,duan2022revisiting} and compare the two-stream\protect\footnotemark\, performance. 
    For NTU-Inter and Volleyball, we only compare the joint modality performance as done in the prior works~\cite{thilakarathne2022pose,zhou2021composer}.
    In Table~\ref{tab-sota-vol}, the method AT$^+$ uses groundtruth bboxes of all video frames for skeleton extraction. }
	\label{tab-sota}
	\vspace{-4mm}
\end{table*}

\subsection{Datasets}

\noindent \textbf{Kinetics-400 (K400)}~\cite{carreira2017quo} contains videos with variable number of people.
Some K400 actions cannot be distinguished by only using skeletons, thus we use a subset of K400, named \textbf{Posetics}~\cite{yang2021unikau}, as the main benchmark for action classification. 
Posetics merges together classes featuring identical skeletons, resulting in 320 total classes (\eg, `eating burger' and `eating hotdog' are merged into `eating something'). 
For K400 and Posetics, we use 2D human skeletons provided by~\cite{duan2022revisiting} (extracted with Faster-RCNN~\cite{ren2015faster} (R50) and HRNet~\cite{sun2019deep}).
We provide K400 results in the appendix Table~\ref{tab-k400}.

\noindent
\textbf{NTU-Inter} is a commonly used benchmark for two-person interaction recognition~\cite{pang2022igformer,perez2021interaction}. 
It's a subset of the in-lab collected NTURGB+D~\cite{liu2020ntu,shahroudy2016ntu}, 
containing only mutual action classes (with two subjects).
Consistent with NTURGB+D, NTU-Inter includes 4 data splits:
NTU60-XSub, NTU60-XView, NTU120-XSub, NTU120-XSet.
We use the official 3D skeletons for NTU-Inter experiments.

\noindent
\textbf{AVA 2.2}~\cite{gu2018ava} is a spatio-temporal action detection benchmark. 
It's a multi-label dataset with 80 action classes, 
falling into three major categories: person movement, object manipulation, and person interaction.
Following the common practice~\cite{wu2019long}, we use Faster-RCNN (ResNeXt-101-FPN~\cite{lin2017feature,xie2017aggregated}) for person detection and HRNet for pose estimation.
The detector is pretrained on COCO and finetuned on AVA training bboxes.
Following the benchmark guidelines, we evaluate 60 action classes. 
We report the overall mAP and the mAP of each major category.

\noindent
\textbf{Volleyball}~\cite{ibrahim2016hierarchical} is a group activity recognition dataset with 8 volleyball group activities. 
Each clip has up to 13 persons, with the center frame annotated with GT person boxes. 
Following common practice~\cite{zhou2021composer,thilakarathne2022pose}, 
we use the tracklets from~\cite{sendo2019heatmapping} and HRNet for pose extraction,
train models on training and validation sets, and report the test set results.

\subsection{Implementation Details}

\noindent
\textbf{Architecture.}
Unless otherwise noted, we adopt ST-GCN++~\cite{duan2022pyskl} as the GCN backbone. 
Particularly, we consider two variants: \textbf{SkeleTR-S} (with fewer GCN blocks and smaller output dimension, details in Table~\ref{tab-arch}), and \textbf{SkeleTR-L} (original ST-GCN++).
The inter-sequence modeling stage consists of a Mix Pooling module and two Transformer encoder blocks.
The feature dimension in each encoder blocks equals to the output channels of the GCN backbone (128 for SkeleTR-S and 256 for SkeleTR-L).

\noindent
\textbf{Training.}
We train all models end-to-end with SGD and batch size 128.
For all models, we set the base learning rate (LR) to 0.1 and adopt a multi-step LR scheduler with the decay factor 0.1.
For Posetics and NTU-Inter, we train models for 100 epochs, and set milestones at the 70$_{th}$ and 85$_{th}$ epoch.
For AVA and Volleyball, we train the models for 40 epochs, and set milestones at the 30$_{th}$ and 35$_{th}$ epoch.
We use cross entropy loss ($\mL_{\mC\mE}$) for Posetics, NTU-Inter and Volleyball, and use binary cross entropy loss ($\mL_{\mB\mC\mE}$) for AVA (multi-label).
We explicitly set a loss weight $\lambda=100$ in all AVA experiments due to the small magnitude of $\mL_{\mB\mC\mE}$.
In Posetics-AVA joint training, we concatenate samples in two datasets and optimize the model with a weighted combination of two losses:
$\mL=\mL_{\mC\mE} + \lambda \mL_{\mB\mC\mE}$. 
For AVA, we sample short skeleton sequences in a small temporal window (5s / 10s) to form a training sample.

\noindent
\textbf{Inference.}
We follow standard practices~\cite{duan2022pyskl} to predict action labels for trimmed videos.
For AVA untrimmed videos, in inference we use a sliding window strategy to generate predictions, with a 1-second time step.
We get an action prediction for a human proposal from each window containing the proposal,
and then average all predictions associated with a human proposal to get the final result.

\subsection{Main Results}

We compare SkeleTR with state-of-the-art methods on three different skeleton action understanding tasks: 
action classification, spatio-temporal action detection, and group activity recognition. 
Unless otherwise noted, all methods use the same skeleton input as SkeleTR.

\footnotetext{Joint and Bone. Each stream takes a separate modality derived from skeleton coordinates as input. 
For joint stream, the input is the coordinates itself. 
For bone stream it is the difference of neighboring joints. }

\noindent
\textbf{Action Classification.}
Table~\ref{tab-sota-pos} compares SkeleTR with state-of-the-art methods on Posetics.
We use official training configurations when available~\cite{liu2020disentangling,duan2022revisiting,yang2021unikau}. 
For methods without official configurations~\cite{chen2021channel,duan2022pyskl},
we use the same hyper-parameter setting as SkeleTR.
SkeleTR with all GCN backbones notably outperforms existing methods.
Compared to CTR-GCN, 
SkeleTR-L achieves 3.6\% higher Top-1 accuracy.
Quantitative analysis shows that the improvement is more significant for multi-person scenarios:
actions with $\ge 5$ persons per frame have around 10\% improvement in mean class Top-1, 
which is twice as much as the improvement for other actions.
SkeleTR also achieves the highest accuracy on Posetics (with OpenPose skeletons) and K400 (with HRNet/OpenPose skeletons).\footnote{Please see Table~\ref{tab-k400} for these results.}
Table~\ref{tab-sota-ntu} shows that on the more specific task of two-person interaction recognition, 
SkeleTR also outperforms existing interaction recognition methods~\cite{perez2021interaction,nguyen2021geomnet,pang2022igformer} and skeleton action classification methods by a large margin.

\noindent
\textbf{Spatio-temporal Action Detection.} 
Skeleton-based spatio-temporal action detection is a new setting first proposed in this paper and we use AVA as the primary benchmark.
For comparisons, we adapt state-of-the-art skeleton action classification algorithms by taking a single skeleton sequence (with 2 seconds length, centered at a person proposal in the key frame) as input to predict the action for this person.
We train all models with the same hyper-parameter setting as SkeleTR.
Due to the lack of global context and interaction modeling, 
all existing methods perform poorly on the new task,
while SkeleTR significantly outperforms them. 
Compared to the baseline ST-GCN++, 
SkeleTR-S improves the overall mAP by 4.1\%.
Among three major action categories, 
SkeleTR-S makes the largest improvement on person interaction (7.0\%), 
thanks to its great inter-person modeling capability.
When jointly trained with Posetics, 
SkeleTR-S enjoys another 4.1\% overall mAP improvement.
However, this time the improvement is basically for person movement and object manipulation, 
since AVA-style person interactions rarely appear in Kinetics videos.

\noindent
\textbf{Group Activity Recognition.}
Most prior works on skeleton-based group activity recognition do not 
utilize the effective GCN architecture to model intra-person motion patterns. 
Instead, they directly model the relationship among skeletons with complicated transformer architectures.
For a comprehensive comparison, we also adapt the state-of-the-art skeleton action recognition algorithms and benchmark them on the Volleyball dataset.
Table~\ref{tab-sota-vol} shows that, with good capability of both intra-sequence motion modeling and inter-sequence relationship modeling, 
SkeleTR outperforms all previous state-of-the-art on Volleyball, including the adapted GCN-based approaches.

\begin{table}[t]
	\captionsetup{font=small}
	\centering 
	\resizebox{\columnwidth}{!}{
    \tablestyle{6pt}{1.2}
    \begin{tabular}{c|cccc}
    \shline
    Methods & Posetics Top-1 & Params (M) & GFLOPs & \makecell{Inference Speed\\(2080Ti, clip/second))} \\ \shline
    ST-TR~\cite{plizzari2021skeleton} & 47.45 & 6.44 & 26.92 & 97 \\ 
    MS-G3D~\cite{liu2020disentangling} & 52.63 & 2.82 & 20.86 & 99  \\ 
    CTR-GCN~\cite{chen2021channel} & 54.10 & 1.36 & 5.36 & 195 \\ 
    ST-GCN++~\cite{duan2022pyskl} & 52.46 & 1.31 & 5.44 & 235 \\ 
    UNIK~\cite{yang2021unikau} & 52.52 & 3.28 & 12.15 & 156 \\ 
    PoseC3D~\cite{duan2022revisiting} & 53.09 & 1.90 & 14.69 & 53 \\ \shline
    SkeleTR-S & 55.85 & \textbf{0.81} & \textbf{2.74} & \textbf{336} \\ 
    SkeleTR-L & \textbf{57.90} & 3.82 & 7.30 & 200 \\ \shline
    \end{tabular}}
    \vspace{-2mm}
    \caption{\textbf{Efficiency statistics of models in Table~\ref{tab-sota-pos}. }}
    \vspace{-2mm}
    \label{tab-cost}
\end{table}

\noindent
\textbf{Computational Efficiency. }
SkeleTR is designed to be efficient and lightweight. 
Table~\ref{tab-cost} shows that SkeleTR-S outperforms previous state-of-the-art methods for skeleton-based action classification in terms of accuracy, while having the model with the smallest parameter size, GFLOPs, and fastest inference speed.
In Table~\ref{tab-efficiency}, we provide additional evidence of SkeleTR's computational efficiency by comparing it to other methods designed for different tasks, such as IGFormer~\cite{pang2022igformer} and COMPOSER~\cite{zhou2021composer}.

\begin{table}[t]
	\captionsetup{font=small}
	\centering 
	\resizebox{\columnwidth}{!}{
	\tablestyle{6pt}{1.3}
    \begin{tabular}{c|cc|cccc}
    \shline
    & \multicolumn{2}{c|}{Posetics} & \multicolumn{4}{c}{AVA v2.2} \\ \shline
    Methods & Top-1 & Top-5 & \makecell{Overall\\mAP} & \makecell{Person\\Movement} & \makecell{Object\\Manipulation} & \makecell{Person\\Interaction} \\ \shline
    RGB & 74.42 & 91.29 & 31.30 & 48.44 & 23.67 & 32.74 \\ \shline
    ST-GCN++ & 52.46 & 75.97 & 18.62 & 39.10 & 9.68 & 19.98 \\ 
    RGB + ST-GCN++ & 75.98 & 91.58 & 31.95 & 52.28 & 23.14 & 33.13 \\ \shline
    SkeleTR & 57.90 & 80.33 & 26.83 & 48.19 & 17.72 & 27.75 \\ 
    RGB + SkeleTR & \textbf{76.70} & \textbf{92.02} & \textbf{34.34} & \textbf{54.78} & \textbf{25.77} & \textbf{34.93} \\ \shline
    \end{tabular}}
    \vspace{-2mm}
    \caption{\textbf{SkeleTR demonstrates great complementarity when fused with RGB recognizers. } We fuse predictions by state-of-the-art RGB recognizers (TimeSformer~\cite{bertasius2021space} for Posetics, ACAR-R101~\cite{pan2021actor} for AVA) with skeleton predictions with $2:1$ ratio. }
    \vspace{-2mm}
    \label{tab-rgb}
\end{table}

\noindent
\textbf{Improved complementarity with RGB modality.}
To demonstrate the complementarity of SkeleTR upon RGB methods,
we directly fuse the prediction of SkeleTR and the state-of-the-art RGB-based action recognizers .
Table~\ref{tab-rgb} shows that late fusion with SkeleTR leads to a significant boost
(2.3\% Posetics top-1, 3\% AVA mAP) compared to the RGB state-of-the-art, 
which notably outperforms ST-GCN++ late fusion.
On AVA, the improvement is extremely large for person movement (6.3\% mAP), 
which emphasizes the importance of the skeleton in modeling human motions.

\subsection{Ablation on Inter-Sequence Modeling}

\begin{table}[t]
    \captionsetup{font=small}
	\centering 
	\resizebox{\linewidth}{!}{
	\tablestyle{10pt}{1.3}
    \begin{tabular}{cccc}
    \shline
    Interaction Modeling & Posetics Top-1 & AVA Overall mAP \\ \shline
    NA & 48.21 & 17.56 \\ \shline
    Add Global & 51.06 & 20.00 \\ 
    Concat Global & 50.01 & 20.14 \\
    Non-Local & 48.43 & 19.78 \\ \shline 
    Transformer Encoders & \textbf{52.81} & \textbf{20.92} \\ \shline
    \end{tabular}}
    \vspace{-2mm}
    \caption{\textbf{Transformer-based interaction modeling significantly outperforms multiple baselines. }}
    \label{tab-abl-baselines}
    \vspace{-2mm}
\end{table}

\begin{table}[t]
    \captionsetup{font=small}
	\centering 
	\resizebox{\linewidth}{!}{
	\tablestyle{6pt}{1.3}
    \begin{tabular}{cccc}
    \shline
    Feature Aggregation & Posetics Top-1 & AVA Overall mAP & \makecell{Head Computation\\(GFLOPs)} \\ \shline
    $1 \times 1$ & 52.81 & 20.92 & 0.013 \\ 
    $T \times P$ & 52.78 & 21.50 & 5.211 \\ \shline
    $T \times 1$ & 53.48 & 20.64 & 0.356 \\ 
    $1 \times P$ & 53.40 & 21.35 & 0.080 \\ 
    $1 \times V$ & 52.15 & 21.01 & 0.429 \\  \shline
    $1 \times 1 + T \times 1$ & 53.75 & 21.00 & 0.392 \\ 
    $1 \times 1 + 1 \times P$ & 53.00 & 21.43 & 0.101 \\ 
    $T \times 1 + 1 \times P$ & 53.16 & 20.87 & 0.550  \\ \shline
    $1 \times 1 + T \times 1 + 1 \times P$ & \textbf{53.90} & \textbf{21.65} & 0.594 \\ \shline
    \end{tabular}}
    \vspace{-2mm}
    \caption{\textbf{Comparing different feature aggregation strategies. }}
    \label{tab-abl-mix}
    \vspace{-3mm}
\end{table}

We conduct ablation experiments to validate the efficacy of our inter-sequence modeling stage. 
In experiments, we adopt the light ST-GCN++ as the backbone. 
We set $M$ to 20, $T$ to 30, and set the temporal window size to 5s for AVA. 
We report the recognition performance with a single joint modality on Posetics and AVA. 

We first compare the Transformer-based modeling with other simple baselines for modeling global context or interaction.
In all experiments, we pool both spatial and temporal dimensions and retain a single feature vector per sequence.
We consider the following baselines: 
1. \textbf{NA}: no interaction modeling;
2. \textbf{Add Global}: process the average feature with an MLP to get a Global feature, then add it back to each sequence feature;
3. \textbf{Concat Global}: concatenate the average feature with each sequence feature, then process with an MLP;
4. \textbf{Non-Local}: process all sequence features with a standard non-local~\cite{wang2018non} block, and apply an MLP to the output.
Table~\ref{tab-abl-baselines} shows that, despite its low computation cost (0.013 GFLOPs / clip),
the Transformer markedly enhances recognition performance relatively to the \textbf{NA} baseline 
(4.6\% Posetics top-1, 3.4\% AVA mAP), 
and considerably outperforms all alternatives.

\begin{figure}[t]
	\centering
	\captionsetup{font=small}
	\includegraphics[width=\columnwidth]{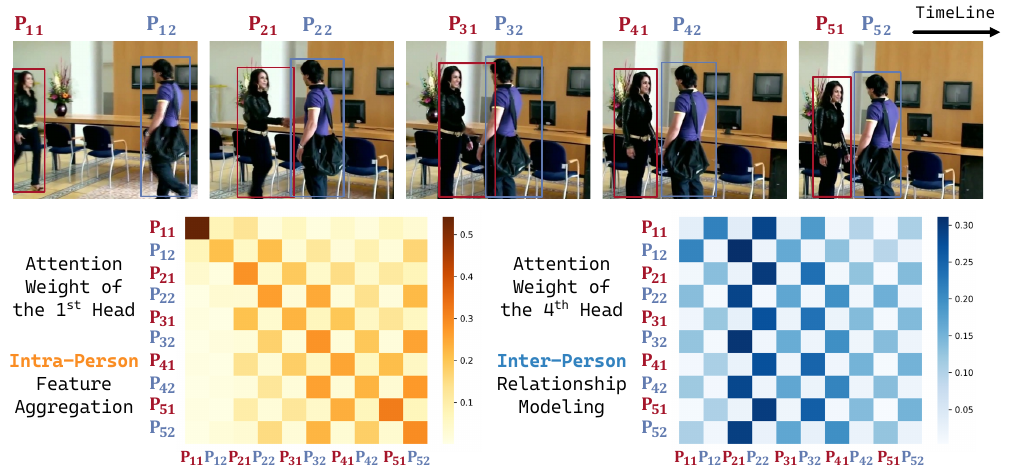}
	\vspace{-5mm}
	\caption{\textbf{Visualization of the weights in different attention heads in the last transformer encoder.}  
	$\mP_{ij}$ denotes the short skeleton sequence of person $j$ at the timestamp $i$. }
	\label{fig-attn}
    \vspace{-1mm}
\end{figure}

Both Non-Local and Transformer encoders adopt self-attention to model the inter-sequence relationship,
while there is a significant difference in recognition performance between them.
The reason is that the multi-head design in Transformers allows for modeling diverse relationships between individual sequences.
Visualizing the attention weights of different attention heads, we find that some of them have clearly interpretable semantics.
Fig~\ref{fig-attn} shows the attention weights of a `handshake' action.
Of the eight attention heads, 
the first head aggregates temporal features within a single person,
while the fourth head models interactions between individuals.
Specifically, the attention weights of the fourth head reveal that $P_{21}$ and $P_{22}$ attend most strongly to other skeleton features, 
which corresponds to the exact moment of the handshaking interaction.

We then delve deeper into the Mix Pooling design, 
which helps to keep fine-grained sequence features and serves as a good trade-off between efficiency and performance.
Table~\ref{tab-abl-mix} compares Mix Pooling ($1\times 1+T\times 1 + 1\times P$) with other feature aggregation strategies.
Mix Pooling outperforms each individual pooling strategy or the combination of two strategies.
It achieves the best performance on both datasets with a reasonable computation cost of 0.594 GFLOPs per clip.

\begin{figure}[t]
	\centering
	\captionsetup{font=small}
	\includegraphics[width=\columnwidth]{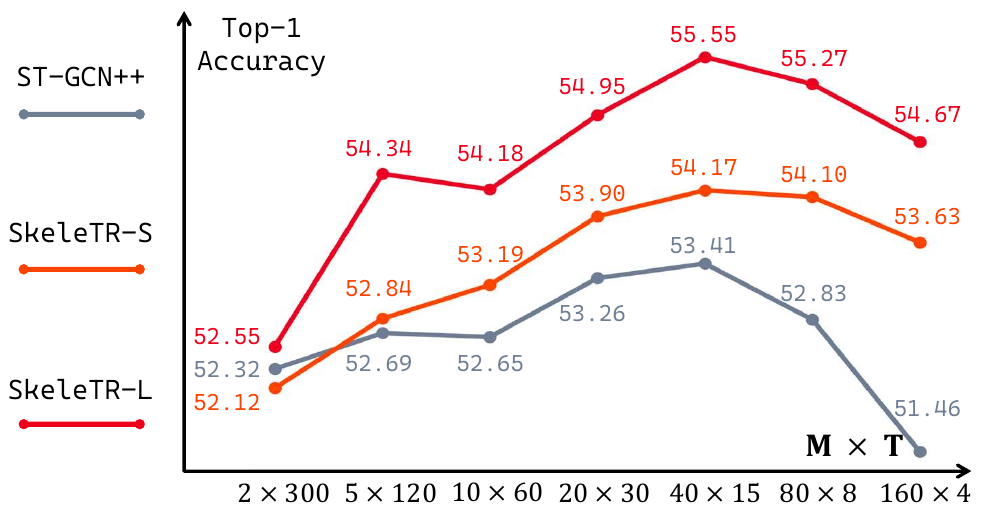}
	\vspace{-5mm}
	\caption{\textbf{Fix the total number of skeletons and adjust the sequence number.} Posetics Top-1 accuracy is visualized.  }
	\label{fig-abl-curve}
    \vspace{-1mm}
\end{figure}

\begin{table}[t]
    \captionsetup{font=small}
	\centering 
	\resizebox{\linewidth}{!}{
	\tablestyle{8pt}{1.3}
    \begin{tabular}{ccccc}
    \shline
    Methods & $M$ & $T$ & Posetics Top-1 & \makecell{Avg GFLOPs for\\Posetics Video Clips} \\ \shline
    SkeleTR-S & $20$ & 30 & 53.90 & 2.430 \\ 
    SkeleTR-L & $[5,20]$ & 30 & \textbf{55.06} & 4.934 \\ 
    SkeleTR-L & $20$ & 30 & 54.95 & 6.653 \\ \shline
    SkeleTR-S & $40$ & 15 & 54.17 & 2.789 \\ 
    SkeleTR-L & $[10,40]$ & 15 & 55.35 & 3.532 \\ 
    SkeleTR-L & $40$ & 15 & \textbf{55.55} & 7.394 \\ \shline
    \end{tabular}}
    \vspace{-2mm}
    \caption{\textbf{Sequence number $M$: fixed or adaptive? } Utilizing an adaptive sequence number $M$ can save great amounts of computational resources and achieve comparable performance. }
    \vspace{-3mm}
    \label{tab-abl-adaptive}
\end{table}

\subsection{Ablation on Input Construction} 
In this section we demonstrate the advantages of our new input format, 
both in recognition performance and computational efficiency. 
We report the recognition performance with a single joint modality.

\noindent
\textbf{Large Number of Short Sequences.}
We use a large number of short skeleton sequences as input for skeleton action understanding tasks.
Besides being a unified input format for different tasks,
the new format also benefits standard skeleton-based action classification under real-world scenarios. 
To investigate the effect of sequence granularity, 
we fix the total number of skeletons $M \times T$, 
and adjust the sequence number $M$.
We compare three models: ST-GCN++, SkeleTR-S, and SkeleTR-L (Fig~\ref{fig-abl-curve}).
Interestingly, we find that all models obtain their highest performance when using a large number of short sequences as input, specifically, $40\times 15$.
Meanwhile, the improvement is especially large for SkeleTR: 
2.1\% top-1 for SkeleTR-S, 3\% top-1 for SkeleTR-L (compared to the $2\times 300$ setting).

\noindent
\textbf{Flexible Number of Sequences.}
SkeleTR is compatible with different numbers of sequences.
To train and test SkeleTR with an adaptive sqeuence number,
we first set a dynamic range $[M_{min}, M_{max}]$,
and for each training batch, 
we randomly sample $\tilde{M}$ from the range and use it as the sequence number for all videos in the batch.
For each testing video, we use the actual sequence number $M'$ if $M_{min} \le M' \le M_{max}$, and cap it to $M_{min}$ or $M_{max}$ otherwise. 
We present an ablation study using two input formats: $20\times 30$ and $40\times 15$. 
For each format, we use a dynamic range ($[5, 20]$, $[10, 40]$) of $M$ to train SkeleTR-L on Posetics, 
and report the results in Table~\ref{tab-abl-adaptive}.
Compared to using a fixed $M$, 
using an adaptive $M$ can achieve comparable recognition performance with a largely reduced inference cost (26\% less GFLOPs for $T=30$, 52\% less for $T=15$). 

\begin{table}[t]
    \captionsetup{font=small}
	\centering 
	\resizebox{\linewidth}{!}{
	\tablestyle{6pt}{1.3}
    \begin{tabular}{cccccc}
    \shline
    Training & \makecell{Temporal\\Window} & \makecell{Overall\\mAP} & \makecell{Person\\Movement\\mAP} & \makecell{Object\\Manipulation\\mAP} & \makecell{Person\\Interaction\\mAP} \\ \shline
    From Scratch & $5$ & 21.65 & 42.34 & 11.37 & 25.66 \\ 
    Finetuning & $5$ & 22.12 & 43.02 & 11.91 & 25.76 \\ 
    Joint Training & $5$ & 23.92 & 44.25 & 14.65 & 26.06 \\ \shline
    From Scratch & $10$ & 21.33 & 42.24 & 11.26 & 24.68  \\ 
    Finetuning & $10$ & 23.11 & 44.06 & 13.32 & 25.85 \\ 
    Joint Training & $10$ & \textbf{25.69} & \textbf{47.13} & \textbf{16.53} & \textbf{26.64}\\ \shline
    \end{tabular}}
    \vspace{-2mm}
    \caption{\textbf{Transfer Learning and Joint Learning with SkeleTR-S. } For a target task with limited training data, an auxiliary dataset can be leveraged to improve the performance and mitigate the overfitting, either by transfer learning or joint learning. }
    \label{tab-abl-unified}
    \vspace{-5mm}
\end{table}

\subsection{Transfer Learning and Joint Learning}

SkeleTR is a general framework for versatile skeleton action recognition tasks, 
which enables transfer learning and joint learning across different tasks.
To demonstrate this, we consider two tasks: 
Posetics action classification and AVA spatio-temporal action detection (with AVA being the smaller dataset).
From-scratch training on AVA is easy to overfit (Fig~\ref{fig-curve}). 
Pretraining on Posetics or joint training on AVA and Posetics both help to mitigate the overfitting, 
with joint training leading to the best recognition performance.
Enlarging the temporal window size for AVA sampling, 
which leads to less diversified training samples, 
amplifies the gain of pretraining or joint training.

\subsection{Limitations}
SkeleTR is currently trained with skeletons pre-extracted from RGB videos, 
and the adopted pose estimation solution is relatively heavy.
To use SkeleTR in real-world apps with limited computation resource, 
additional efforts are required to switch to a light-weight pose estimation solution that can perform fast inference on CPUs or edge devices, and to retrain the models accordingly.
Besides, the short sequence format proposed by SkeleTR strikes a good balance between mitigating skeleton matching noises and modeling intra-sequence dynamics in real-world scenarios.
The format may not lead to extra benefits when the ground-truth skeleton sequences are available (Table~\ref{tab-limit}).

\section{Conclusion}

We presented SkeleTR, a unified framework for skeleton action recognition in general scenarios.
It extends existing GCN backbones and achieves state-of-the-art performance on multiple skeleton action understanding tasks, including video-level, instance-level, and group-level action recognition.
Besides, SkeleTR is compatible with transfer learning and joint learning across different datasets and tasks, 
which further improves the capability of learned representations.

\noindent \textbf{Broader Impact.}
Using human skeletons for action recognition is more robust to environment changes and can help preserve user privacy.
Real-world applications of skeleton action recognition include monitoring patient or elderly health, enabling autonomous driving, and facilitating human-computer interactions.
However, there could be unintended usages and we advocate responsible usage complying with applicable laws and regulations.

\appendix

\section{Implementation Details}

\begin{table}[t]
    \captionsetup{font=small}
	\centering 
	\resizebox{\linewidth}{!}{
	\tablestyle{4pt}{1.3}
    \begin{tabular}{cccc}
    \shline
    \multicolumn{4}{c}{Configuration}                                               \\ \shline
    Backbone & ST-GCN++[S] & ST-GCN++~\cite{duan2022pyskl} & CTR-GCN~\cite{chen2021channel} \\ \shline
    Num Stages                     & 6                & 10       & 10  \\
    Base Channel Width             & 64               & 64       & 64  \\
    Channel Inflation Stages       & $6_{th}$ & $5_{th}, 8_{th}$ & $5_{th}, 8_{th}$  \\
    Temporal DownSample Stages     & $6_{th}$ & $5_{th}, 8_{th}$ & $5_{th}, 8_{th}$  \\
    Output Channel Width           & 128              & 256      & 256 \\ \shline
    \multicolumn{4}{c}{Statistics}                                                  \\ \shline
    Parameter Size (M)             & 0.181            & 1.314    & 1.356  \\
    FLOPs for a T=300 Sequence (G) & 0.920            & 2.766    & 2.717  \\
    FLOPs for a T=30 Sequence (G)  & 0.092            & 0.284    & 0.298  \\ \shline
    \end{tabular}}
    \caption{\textbf{The detailed architectures and statistics of the GCN backbones used in SkeleTR. }}
    \label{tab-arch}
    \vspace{-3mm}
\end{table}

\noindent
\textbf{GCN backbone Architecture.} 
Table~\ref{tab-arch} describes the detailed architectures of the GCN backbones used in our work. 
ST-GCN++[S], in contrast to STGCN++~\cite{duan2022pyskl}, has fewer GCN blocks and a smaller output feature dimension, resulting in a lighter version. 
Other GCN backbones maintain the same architecture as their corresponding papers.
The statistics show that ST-GCN++[S] is a very light GCN backbone. 
It has the parameter size reduced to $1/7$ and the computation cost for each skeleton sequence reduced to $1/3$ compared to ST-GCN++.

\begin{figure}[t]
	\centering
	\captionsetup{font=small}
	\includegraphics[width=.5\columnwidth]{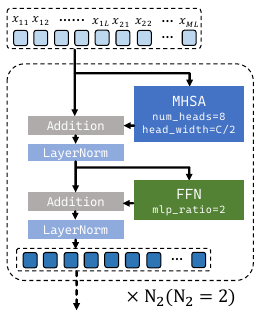}
	\caption{\textbf{The architecture of the Transformer encoders. } }
	\label{fig-encoder}
	\vspace{-3mm}
\end{figure}

\noindent
\textbf{Transformer Architecture.}~Fig~\ref{fig-encoder} shows the architecture of Transformer encoders used in SkeleTR in detail.
The Transformer is a stack of two encoder blocks. 
Each encoder block adopts the post-layernorm style~\cite{xiong2020layer} and consists of a multi-head self-attention layer (MHSA) and a feed-forward network (FFN). 
We define $C$ as the input channel width of the Transformer. 
The MHSA has 8 attention heads, with the channel width of each attention head set to $C / 2$. 
The FFN is instantiated with an MLP with two fully connected layers, with the input and output dimension set to $C$, and the hidden dimension set to $2C$. 
For consistency, we use the above hyper parameters in all our experiments. 
Tuning these hyper parameters may lead to additional performance improvement,
but it was not the focus of our work.

\section{Supplementary Experiments}

\subsection{Main Results}

\begin{table*}[t]
    \captionsetup{font=small}
	\centering 
	\resizebox{\linewidth}{!}{
	\tablestyle{6pt}{1.3}
    \begin{tabular}{c|cc|cc|cc|cc}
    \shline
    Methods & \multicolumn{2}{c|}{Posetics + HRNet Skeleton} &  \multicolumn{2}{c|}{Posetics + OpenPose Skeleton} & 
    \multicolumn{2}{c|}{Kinetics-400 + HRNet Skeleton} &  \multicolumn{2}{c}{Kinetics-400 + OpenPose Skeleton} \\ \shline
    & Joint Only & Joint + Bone & Joint Only & Joint + Bone & Joint Only & Joint + Bone & Joint Only & Joint + Bone \\ \shline
    \makecell{The State-of-the-art in\\Previous Works} & / & / & / & \makecell{45.9\\(UNIK~\cite{yang2021unikau})} 
    & \makecell{46.0\\(PoseConv3D~\cite{duan2022revisiting})} & \makecell{47.7\\(PoseConv3D~\cite{duan2022revisiting})}  
    & \makecell{36.6\\(DualHead~\cite{chen2021learning})} & \makecell{38.3\\(DualHead~\cite{chen2021learning})} \\ \shline
    ST-GCN++~\cite{duan2022pyskl} & 50.99 & 52.46 & 41.18 & 43.64 & 42.78 & 44.29 & 31.18 & 33.13 \\ 
    SkeleTR-L & 55.55 \inc{4.56} & 57.90 \inc{5.44} & 45.90 \inc{4.72} & 48.02 \inc{4.38} & 
    47.79 \inc{5.01} & 49.36 \inc{5.07} & 37.35 \inc{6.17} & 39.43 \inc{6.30} \\ \shline
    CTR-GCN~\cite{chen2021channel} & 51.43 & 54.10 & 41.22 & 44.78 & 44.65 & 46.10 & 33.21 & 35.03 \\ 
    SkeleTR (CTR-GCN) & 55.46 \inc{4.03} & 57.56 \inc{3.46} & 46.12 \inc{4.90} & 48.84 \inc{4.06} &
    47.86 \inc{3.21} & 49.54 \inc{3.34} & 36.91 \inc{3.70} & 39.29 \inc{4.26} \\ \shline
    \end{tabular}}
    \caption{\textbf{Skeleton-based action classification results on four Kinetics-400 benchmarks. } 
    We evaluate SkeleTR on both the Posetics~\cite{yang2021unikau} split and the original Kinetics-400~\cite{yan2018spatial} split, with OpenPose~\cite{8765346} 2D skeletons or HRNet~\cite{sun2019deep} 2D skeletons. 
    SkeleTR achieves the best accuracy on all four benchmarks and significantly outperforms the state-of-the-art in previous works. }
    \label{tab-k400}
    \vspace{-3mm}
\end{table*}

\noindent
\textbf{Results on more Kinetics-400 splits.}
To demonstrate the generality of SkeleTR, 
we instantiate it with two different GCN backbones (ST-GCN++~\cite{duan2022pyskl}, CTR-GCN~\cite{chen2021channel}) and evaluate on four different Kinetics-400 benchmarks. 
We adopt $M=40$ and $T=15$ as the input format and report the accuracy of the joint stream and the joint+bone late fusion.
As shown in Table~\ref{tab-k400}, on all four benchmarks, 
SkeleTR achieves the best accuracy and outperforms state-of-the-art methods significantly.
Moreover, the SkeleTR framework largely improves the recognition performance compared to the original GCN methods, achieving 3\% to 6\% higher top-1 accuracy on all the backbones and benchmarks.

\begin{table}[t]
	\captionsetup{font=small}
	\centering 
	\resizebox{\columnwidth}{!}{
    \tablestyle{6pt}{1.2}
    \begin{tabular}{c|ccc}
    \shline
    Methods & NTU120-XSub-Inter Top-1 & Params (M) & GFLOPs \\ \shline
    IGFormer~\cite{pang2022igformer} & 85.4 & 20.29$^+$ & 5.08$^+$ \\ 
    SkeleTR-S & \textbf{87.4} & \textbf{0.81} & \textbf{2.74} \\ \shline
    Methods & Volleyball Top-1 & Params (M) & GFLOPs \\ \shline
    COMPOSER~\cite{zhou2021composer} & 93.7 & 10.59 & 0.72 \\ 
    SkeleTR-S & \textbf{94.4} & \textbf{0.81} & \textbf{0.53} \\ \shline
    \end{tabular}}
    \vspace{-3mm}
    \caption{\textbf{Comparing the efficiency of SkeleTR with IGFormer and COMPOSER. }
    There is no publicly available IGFormer implementation.
    As a result, we use the implementation details described in the original paper~\cite{pang2022igformer} to estimate the parameter size and FLOPs of IGFormer.
    $^+$ denotes that the real parameter size and FLOPs are larger than our estimation, 
    since our estimation is limited to a specific component (Interaction Transformer Blocks) within IGFormer.}
    \label{tab-efficiency}
\end{table}

\noindent 
\textbf{Computational Efficiency. }
We compare the computational efficiency of SkeleTR with the latest methods for two-person interaction recognition and group activity recognition.
We excluded methods without publicly available source codes or sufficient implementation details and include two methods: IGFormer~\cite{pang2022igformer} and COMPOSER~\cite{zhou2021composer}.
Table~\ref{tab-efficiency} shows that SkeleTR-S is more efficient than the alternatives.
Specifically, SkeleTR has significantly fewer parameters than other methods: $25\times$ less than IGFormer and $13\times$ less than COMPOSER.

\subsection{Quantitative Analysis}

\begin{table}[t]
    \captionsetup{font=small}
	\centering 
	\resizebox{\linewidth}{!}{
	\tablestyle{8pt}{1.3}
    \begin{tabular}{ccc}
    \shline
    Action & Top-1 Improvement & Average $\mathrm{N_{ske}}$ \\ \shline
    Swing Dancing & $38.5\rightarrow73.1$ \inc{34.6} & 7.2 \\
    Shaking Hands & $11.4\rightarrow42.9$ \inc{31.5} & 4.1 \\
    Dodgeball & $37.5\rightarrow67.5$ \inc{30.0} & 9.1 \\
    Cheerleading & $48.4\rightarrow77.4$ \inc{29.0} & 15.2 \\
    Playing Volleyball & $45.2\rightarrow69.0$ \inc{23.8} & 9.0 \\ \shline
    \end{tabular}}
    \caption{\textbf{Actions with top-5 largest accuracy improvement in Posetics. } 
    We compare the Top-1 classification accuracy (joint+bone) of ST-GCN++ (left) and SkeleTR-L (right) at the action class level, 
    and also list the improvement. 
    Both models are trained from-scratch on Posetics.}
    \label{tab-topk}
    \vspace{-3mm}
\end{table}

\noindent
\textbf{Quantitative Analysis on Posetics.}~SkeleTR can effectively leverage all skeletons in a video for action classification. 
In the main paper, we mention that for Posetics action classification, 
the improvement brought by SkeleTR is more significant for videos with lots of people.
For each video, we define $\mathrm{N_{ske}}$ as the average number of skeletons per frame\footnote{
Only frames with human skeletons count when calculating the average. So $\mathrm{N_{ske}}$ will be at least 1.}.
Specifically, the Top-1 accuracy improvement of SkeleTR-L (compared to ST-GCN++) is 3.2\% for videos with $1\le\mathrm{N_{ske}}<2$, 
5.4\% for videos with $2\le\mathrm{N_{ske}}<5$, and 13.1\% for videos with $5\le\mathrm{N_{ske}}$.
We list the actions with top-5 largest accuracy improvement in Posetics in Table~\ref{tab-topk}.
We find that, on average, videos belong to these action categories have a larger average $\mathrm{N_{ske}}$ 
compared to the average $\mathrm{N_{ske}}$ of all validation videos (which is 3.2). 

\begin{table}[t]
    \captionsetup{font=small}
	\centering 
	\resizebox{\linewidth}{!}{
	\tablestyle{4pt}{1.3}
    \begin{tabular}{ccc}
    \shline
    Action & mAP Improvement & Category \\ \shline
    Hand Clap & $24.8\rightarrow44.8$ \inc{20.0} & Person Interaction \\
    Listen to (a person) & $49.8\rightarrow65.9$ \inc{16.1} & Person Interaction \\
    Ride (e.g., a bike, a car, a horse) & $28.7\rightarrow41.8$ \inc{13.0} & Object Manipulation \\
    Hug (a person) & $9.9\rightarrow21.0$ \inc{11.1} & Person Interaction \\
    Lie / Sleep & $37.8\rightarrow48.9$ \inc{11.1} & Person Movement \\ \shline
    \end{tabular}}
    \caption{\textbf{Actions with top-5 largest accuracy improvement in AVA v2.2. }
    We compare the class mAP (joint+bone) of ST-GCN++ (left) and SkeleTR-S (right), and also list the improvement. 
    Both models are trained from-scratch on AVA v2.2.}
    \label{tab-topk-ava}
    \vspace{-3mm}
\end{table}

\noindent
\textbf{Quantitative Analysis on AVA. }
In the main paper, we show that SkeleTR makes the most significant improvement on person interaction. 
In Table~\ref{tab-topk-ava}, we list AVA actions with the largest improvement in action detection performance. 
Among five actions, three of them belong to the person interaction category, 
while the other two actions are about single-person movements\footnote{
Though `Ride (e.g., a bike, a car, a horse)' is officially categorized as `Object Manipulation'. 
After looking into multiple samples, we find that in many cases they can be recognized solely based on person movements. }.
Such improvement is due to the improved capability to model inter-person relationship (which improves person interaction)
and the long-range context (which improves person movement). 

\begin{figure*}[t]
	\centering
	\captionsetup{font=small}
	\includegraphics[width=\linewidth]{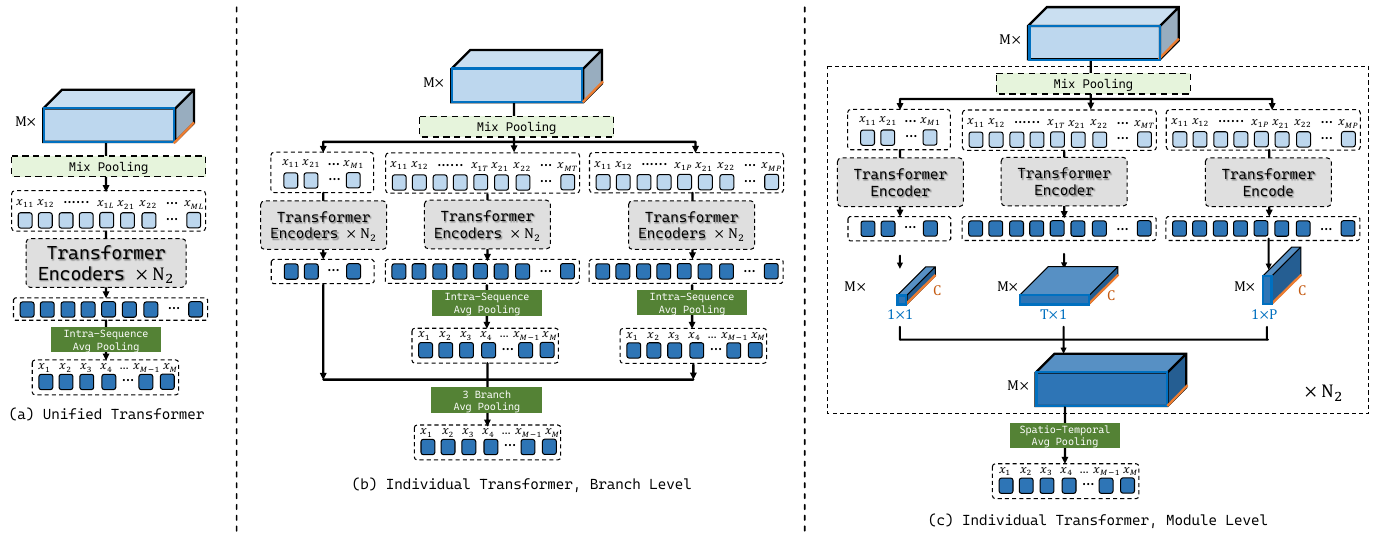}
	\caption{\textbf{Three designs for modeling the multi-view skeleton features. } 
	(a) Using a unified Transformer to model the multi-view skeleton features. 
	(b) Using a separate Transformer to model skeleton features from each pooling branch, and do late fusion at the end. 
	(c) Using a separate Transformer encoder to model skeleton features from each branch, reshape the output of the encoders and add them back (with broadcasting) to form spatio-temporal skeleton features again. Repeat that $\mathrm{N_2}$ times. 
	Among three designs, design (a) achieves the best recognition performance and has the smallest parameter size, thus is adopted in SkeleTR. }
	\label{fig-alter}
	\vspace{-3mm}
\end{figure*}

\subsection{Ablation Study}

Unless otherwise specified, in all ablation experiments, 
we use the SkeleTR-S as our base model with input configuration $M=20$ and $T=30$.  

\begin{table}[t]
    \captionsetup{font=small}
	\centering 
	\resizebox{\linewidth}{!}{
	\tablestyle{6pt}{1.3}
    \begin{tabular}{cccc}
    \shline
    Methods & Sequence Matching & Posetics Top-1 & Posetics Top-5 \\ \shline
    ST-GCN++ & score-based & 50.11 & 74.58  \\ 
    ST-GCN++ & iou-based & \textbf{52.32} & \textbf{75.57} \\  \shline
    SkeleTR-S & score-based & 48.21 & 73.16 \\ 
    SkeleTR-S & iou-based & \textbf{52.12} & \textbf{75.92} \\ \shline
    \end{tabular}}
    \caption{\textbf{IoU-based matching consistently outperforms score-based one for different architectures. } We use the input format $M=2, T=300$ for both ST-GCN++ and SkeleTR. }
    \label{tab-abl-matching}
    \vspace{-3mm}
\end{table}

\noindent
\textbf{IoU-based Matching.}
We use IoU-based matching to link skeletons into sequences.
Compared to the score-based matching in the pre-processing stage, 
our matching strategy not only helps to generate sequences with better consistency, 
but also avoids the irreversible information loss in pre-processing.
To present a direct comparison,
we compare two matching strategies using the input format adopted in most previous works, $M=2,T=300$.

\begin{table}[t]
    \captionsetup{font=small}
	\centering 
	\resizebox{\linewidth}{!}{
	\tablestyle{6pt}{1.3}
    \begin{tabular}{cccc}
    \shline
    Strategy & Posetics Top-1 & \makecell{Head Computation\\(GFLOPs)} & \makecell{Head Parameters\\(MParams)} \\ \shline
    \makecell{Mix Pooling with\\Individual Transformer (B)} & 52.47 & 0.449 & 1.880 \\ 
    \makecell{Mix Pooling with\\Individual Transformer (M)} & 52.32 & 0.449 & 1.880 \\ \shline
    \makecell{Mix Pooling with\\Unified Transformer} & 53.90 & 0.594 & 0.627 \\ \shline
    \end{tabular}}
    \caption{\textbf{Comparing different designs for modeling the multi-view skeleton features. }
    (B) means using separate Transformers at the \textbf{B}ranch level, corresponds to Fig~\ref{fig-alter}(b); 
    (M) means using separate Transformers at the \textbf{M}odule level, corresponds to Fig~\ref{fig-alter}(c).}
    \label{tab-abl-trdesign}
    \vspace{-3mm}
\end{table}

\noindent
\textbf{The disadvantage of using separate Transformers for each feature aggregation stream.} In the inter-sequence modeling stage, we use a single Transformer to model the multi-view skeleton features. 
In experiments, we have also compared our strategy to two alternative designs depicted in Fig~\ref{fig-alter}.
Table~\ref{tab-abl-trdesign} shows that, 
compared to using separate Transformers for each single feature aggregation stream, 
using a unified Transformer leads to better recognition performance with far fewer parameters.

\begin{table}[t]
    \captionsetup{font=small}
	\centering 
	\resizebox{.8\linewidth}{!}{
	\tablestyle{6pt}{1.2}
    \begin{tabular}{ccc}
    \shline
    Intra-Sequence Modeling & Posetics Top-1 & AVA mAP \\ \shline
    \makecell{MLP} & 45.19 & 15.96 \\
    \makecell{Transformer (ST-TR~\cite{plizzari2021skeleton})} & 51.35 & 19.92 \\ 
    \makecell{GCN (ST-GCN++[S])} & 53.90 & 21.65  \\ \shline
    \end{tabular}}
    \caption{\textbf{The GCN significantly outperforms other architectures in intra-sequence dynamics modeling. } 
    Here we take ST-GCN++[S] as the example, while other GCN backbones (\textit{e.g.}, CTR-GCN) also share such advantages.}
    \label{tab-abl-stage1}
    \vspace{-3mm}
\end{table}

\noindent
\textbf{Using different architectures for intra-sequence modeling.}
SkeleTR adopts the GCN architecture to model the motion patterns of each short skeleton sequence, 
considering their superior intra-sequence modeling capability.
In Table~\ref{tab-abl-stage1}, we compare the GCN to two alternative architectures:
(1) \textbf{MLP}: resize skeleton sequences $\mX_i \in \mathbb{R}^{T\times V\times C}$ to vectors, and process them with a two-layer MLP (with hidden dimension 256); 
(2) \textbf{Transformer}: use Transformer-based architectures (specifically, we use ST-TR~\cite{plizzari2021skeleton}) for intra-sequence modeling. 
Experiment results show that replacing the GCN with other architectures in the intra-sequence modeling stage severely undermines the action recognition performance.

\begin{figure*}[t]
	\centering
	\captionsetup{font=small}
	\includegraphics[width=\linewidth]{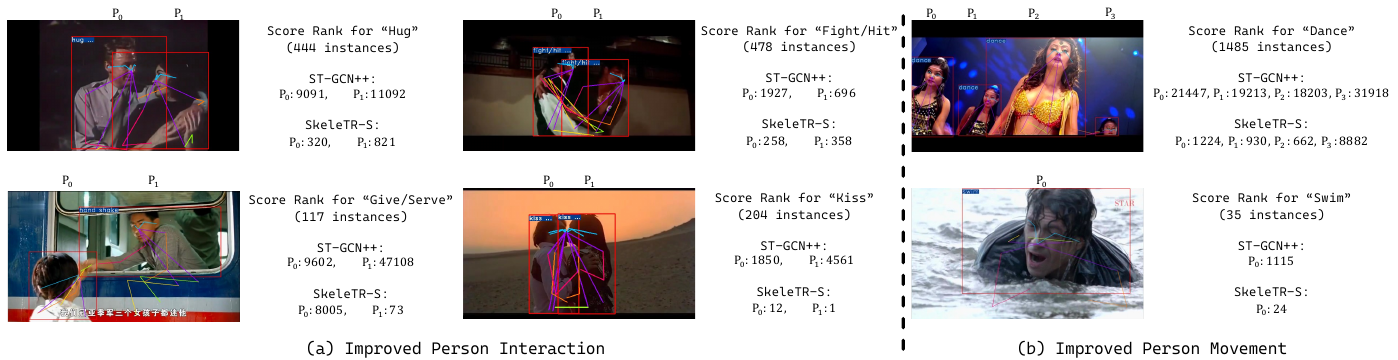}
	\caption{\textbf{Visualization of the action detection results on AVA v2.2 samples (predicted by ST-GCN++ and SkeleTR-S). } 
	For each sample, we list the prediction results for a specific action. 
	The blue label-box at the top-left corner of the proposal indicates that a person is performing the labeled action. 
	Since AVA is a highly imbalanced multi-label dataset, instead of the action score, we show the score rank of one proposal among all proposals in AVA validation set ($\sim$ 148K).  
	The smaller the rank, the larger the model confidence that the person is performing the corresponding action. }
	\label{fig-quali}
	\vspace{-3mm}
\end{figure*}

\subsection{Qualitative Results} 

Compared to ST-GCN++, SkeleTR significantly improves the mAP of Person Interaction and Person Movement actions.
In Fig~\ref{fig-quali}, we visualize some samples in AVA v2.2 validation set.
For each human proposal, we report the relative score rank of a specific action among the corresponding scores of all human proposals ($\sim$ 148K). 
Fig~\ref{fig-quali} shows that, 
for actors performing a Person Interaction or Person Movement action, 
the score rank predicted by SkeleTR is much smaller than the score rank predicted ST-GCN++. 
That demonstrates the superior capability of SkeleTR on recognizing actions that belong to these two major categories. 

\subsection{Limitations}

\noindent 
\textbf{Inputting short skeleton sequences does not improve performance when GT long skeleton sequences present.}
Ensuring that all skeletons in a long input sequence refer to the same identity is challenging due to long-term tracking difficulties.
Prior works mainly focus on simplified datasets such as the standard NTU-RGBD (with T=300), which involves at most two actors and provides ground-truth skeleton sequences, making this issue less of a concern.
In our work, we tackle more complex datasets that involve a multitude of people, which poses challenges in obtaining long sequences free of errors.
SkeleTR mitigates this problem by using very short sequences, and have the model intrinsically learn how sequences and actors interact with each other. 
Table~\ref{tab-limit} shows that although using shorter inputs does not improve performance when long ground-truth sequences are available, it is essential for yielding better performance in general scenarios.

\begin{table}[t]
    \captionsetup{font=small}
	\centering 
	\resizebox{\linewidth}{!}{
	\tablestyle{6pt}{1.2}
    \begin{tabular}{c|c|ccccc}
        \shline
        \multirow{2}{*}{Dataset} & \multirow{2}{*}{Method} & \multicolumn{5}{c}{$M \times T$ ($M$: number of sequence; $T$: sequence length)} \\
        & & $2\times300$ & $10\times 60$ & $20\times 30$ & $40\times 15$ & $80\times 8$ \\ \shline
        \multirow{2}{*}{\makecell{NTU60-XSub \\ (controlled env)}} & ST-GCN++ & \textbf{88.8} & 87.9 & 86.6 & 85.9 & 85.9 \\ 
        & SkeleTR-L & \textbf{89.0} & 88.5 & 87.5 & 87.2 & 86.2 \\ \shline
        \multirow{2}{*}{\makecell{Posetics \\ (real-world env)}} & ST-GCN++ & 52.3 & 52.7 & 53.3 &  \textbf{53.4} & 52.8 \\ 
        & SkeleTR-L & 52.6 & 54.3 & 55.0 & \textbf{55.6} & 55.3 \\ \shline
        \end{tabular}}
    \vspace{-3mm}
    \caption{\textbf{Comparing the effect of using an increasing number of short sequences for controlled environment and real-world scenarios.}}
    \label{tab-limit}
    \vspace{-5mm}
\end{table}

{\small
\bibliographystyle{ieee_fullname}
\bibliography{egbib}
}

\end{document}